\documentclass[11pt]{article}
\usepackage{coling2020}
\usepackage{times}
\usepackage{url}
\usepackage{latexsym}
\usepackage{graphicx}
\usepackage[fleqn]{amsmath}
\usepackage{booktabs}
\usepackage{multirow}
\usepackage{amssymb}
\usepackage{float}

\title{Deep Bag-of-Sub-Emotions for Depression Detection in Social Media}

\colingfinalcopy

\author{Juan S. Lara, Mario Ezra Arag\'on, Fabio A. Gonz\'alez, Manuel Montes-y-G\'omez\\}

\begin{document}
\maketitle

\begin{abstract}
  This paper presents the Deep Bag-of-Sub-Emotions (DeepBoSE), a novel deep learning model for depression detection in social media. The model is formulated such that it internally computes a differentiable Bag-of-Features (BoF) representation that incorporates emotional information. This is achieved by a reinterpretation of classical weighting schemes like term frequency-inverse document frequency into probabilistic deep learning operations. An important advantage of the proposed method is that it can be trained under the transfer learning paradigm, which is useful to enhance conventional BoF models that cannot be directly integrated into deep learning architectures. Experiments were performed in the eRisk17 and eRisk18 datasets for the depression detection task; results show that DeepBoSE outperforms conventional BoF representations and it is competitive with the state of the art, achieving a F1-score over the positive class of 0.64 in eRisk17 and 0.65 in eRisk18.
\end{abstract}
\section{Introduction}

Nowadays, millions of people around the world are affected by different mental disorders that interfere in their thinking and behavior, damaging their quality of life \cite{depression08,depression09}. Timely detection of mental disorders is important to help people before the illness gets worse, minimizing disabilities and returning them to their normal life. Depression is one of the most common mental disorders and a leading cause of risk for suicide \cite{depression08}. It is a serious medical condition associated with loss of interest, a significant change in weight or appetite, and insomnia. It is a highly debilitating condition and a major cause of disability. Depression affects millions of people worldwide, no matter the culture, gender, age, race, or economic status. Currently, only around 20\% of the affected people receive treatment, and most of the spending on mental health is used to maintain psychiatric hospitals instead of detection, prevention, and rehabilitation \cite{mhMexico}. Based on these facts, it is imperative to create effective methods to detect depression before it causes irreparable damage to people affected by this disease.

In another order of ideas, it is a fact that for many people the majority of their social life does not take place in their immediate environment, but in a virtual world created by social media platforms like Facebook, Twitter, Reddit, Instagram, among others. This presents a great opportunity to understand depression through the analysis of social media documents, increasing the chances to detect people that present signs of depression, and providing professional assistance as soon as possible \cite{guntuku2017detecting,depression001}. In this matter, several natural language processing (NLP) methods have been used for depression detection \cite{depression07}, especially, linguistic and sentiment analysis are applied to determine the posts' polarity, and to represent the users by histograms of the ratios of their positive, negative and neutral posts.

Although conventional NLP approaches provide a good measure of the emotions in the text data, recent works in depression detection \cite{subEmo} have demonstrated that a better performance is achieved using fine-grained representations. For instance, the Bag-of-Sub-Emotions (BoSE) is a representation that creates a dictionary of sub-emotions using a clustering strategy and a lexical resource of emotions. Each word in the users' posts is replaced with a label of its closest sub-emotion and a histogram of the sub-emotions is computed as the final representation. An important advantage of BoSE is that it achieves a very good performance while preserving interpretability, which differs from most of the state-of-the-art methods that use deep learning for improved performance but cannot be easily interpreted. The main disadvantage of BoSE is that it relies on feature engineering, i.e., it separates the representation and the prediction phases, because, a Bag-of-Features (BoF) must be offline learned and a classifier must be separately trained. In this concern, an end-to-end neural network model has the potential advantage of combining the representation and the classification phases in a fully trainable model that integrates the expressive power of the BoF representation while making it possible to fine-tune it to get higher performance.

Based on the above ideas, we present a novel deep learning model that internally computes an interpretable BoSE representation while taking advantage of deep representation learning. This is achieved using a differentiable reformulation of a BoF representation that is incorporated into an end-to-end model. The main contributions of this work are:

\begin{itemize}
      \item The Deep Bag of Sub-Emotions (DeepBoSE) model for depression detection that extends the BoSE representation using probabilistic deep learning components.
      \item A training strategy that combines unsupervised and supervised learning. On the one hand, unsupervised information is used to enhance the clustering strategy that defines the sub-emotions; on the other hand, supervised information is incorporated to enhance the representation.
\end{itemize}
\section{Related Work}

The assessment of social media data is an application that mainly concerns NLP and sentiment analysis; it is a promising alternative for the automatic diagnostic of mental illness \cite{guntuku2017detecting}. For instance, it has been successfully used for the automatic assessment of several mental health conditions including borderline personality disorder, bipolar disorder, schizophrenia, anxiety, depression, self-harming, suicidal tendency, addiction, alcoholism, autism, among others \cite{gkotsis2017characterisation}. These applications have gained special attention in recent years, considering that an automated diagnostic system can be potentially applied for early detection \cite{guntuku2017detecting}. Likewise, the development of new methods is also encouraged by popular competitions like the early risk prediction on the internet (eRisk) that is annually performed in the CLEF conference \cite{erisk2018}. Several text processing strategies have been explored for depression detection; they can be divided in two main approaches:

\textit{Feature engineering-based approaches}: the fundamental component of these kinds of methods is the design of a meaningful representation that captures the general topics and sentiments that users express in their posts. More precisely, a descriptor of a document is computed and classification models like support vector machines, logistic regression, or random forests are used to predict a depression label \cite{guntuku2017detecting}. Examples of the most relevant used features include word and character ngrams with TF-IDF weighting \cite{coppersmith2016exploratory}, topic probabilities estimated through latent Dirichlet allocation \cite{yazdavar2017semi}, statistical features, part-of-speech labels \cite{sawhney2018computational}, linguistic structure measures, and interpersonal awareness and interaction \cite{de2016discovering}.

Although these general NLP descriptors provide a good overall representation of the documents, there are domain-specific features that better describe the emotional and specialized content in the posts, and therefore that are more appropriate for the depression detection task. For example, there is evidence that hypo-maniac states, levels of guilt or shame, and the presence of emotions like anger or sadness in posts of depressed users are correlated with suicide attempts \cite{coppersmith2016exploratory}. In order to capture this, a common strategy is to design descriptors that, for example, measure the emotional valence or polarity in the posts \cite{de2014characterizing,reece2017forecasting}. Similarly, detailed descriptors like Bag-of-Emotions (BoE) or specialized Bag-of-Features are used to represent the ratios of terms associated with certain emotions or categories like pronouns, cognitive processes, health, among others \cite{eichstaedt2018facebook}. In this matter, a common approach is to use the Linguistic Inquiry and Word Count (LIWC) \cite{liwc1}, which determines the most psychologically meaningful words according to categories like social relationships, thinking styles or individual differences. LIWC also includes additional lexicons that are typically used to compute these detailed representations \cite{reece2017forecasting,de2014characterizing,eichstaedt2018facebook,sawhney2018computational}. More recently, it was proposed an interesting alternative that consists of using a fine-grained representation of the emotions, referred to as Bag of Sub-Emotions \cite{subEmo}. This idea extends the BoE representations, specifically, the terms are not only assigned to a specific emotion, instead, they are assigned to sub-groups that provide a better approximation of the distribution of the emotions in the texts.

\textit{Deep learning-based approaches}: recently, deep learning has been used as an alternative to classical feature engineering approaches. Neural networks can automatically learn a representation through several non-linear transformations, which is useful considering that it does not require the effort and the domain knowledge for the manual design of the features. An important advantage of deep representation learning is that it can incorporate supervised information to enhance the descriptors, obtaining specialized representations for depression detection. The most remarkable examples of deep learning applied to depression detection include convolutional neural networks \cite{orabi2018deep}, recurrent neural networks \cite{orabi2018deep,gkotsis2017characterisation}, and approaches that use special components like attention mechanisms \cite{cong2018xa}.

There is a trade-off between feature engineering and deep learning approaches. On the one hand, feature engineering provides methods that are interpretable, and, therefore, that are useful to discover patterns in the data that may help humans to distinguish between depressed and healthy users. However, these methods require an offline computation of the representation, which restricts them to be static during the training of the classification model and to achieve non-outstanding performances. On the other hand, the deep learning approaches automatically learn an appropriate descriptor and combines the representation and classification in a single model, allowing to adapt the representations to achieve higher performances. However, the main disadvantage is that the learned representations in a neural network cannot be easily interpreted. We formulate the DeepBoSE model such that it takes advantage of the interpretability of BoSE as well as from the representation learning capabilities of neural networks. As it will be detailed in the next section, the proposed method extends BoSE using probabilistic notions that are embedded in a deep neural network that allows transfer learning.
\section{Deep Bag-of-Sub-Emotions}

To introduce the Deep Bag of Sub-Emotions (DeepBose), we firstly summarize the BoSE approach, which consists of three main steps: first, a set of fine-grained emotions are unsupervisedly learned from a lexical resource that contains words associated to different emotions and sentiments, this is achieved using a clustering technique that discretizes the distribution of each emotion $e$ in $K_e$ sub-groups (named as sub-emotions). Second, the fine-grained emotions are used to represent the documents, specifically, each word is masked or substitute by its closer sub-emotion, and each document is represented by a frequency histogram of their sub-emotions. Third, the histogram representation is used to train a classification model that predicts the depression label. 

\begin{figure}[t!]
	\begin{center}
		\includegraphics[width=0.75\columnwidth]{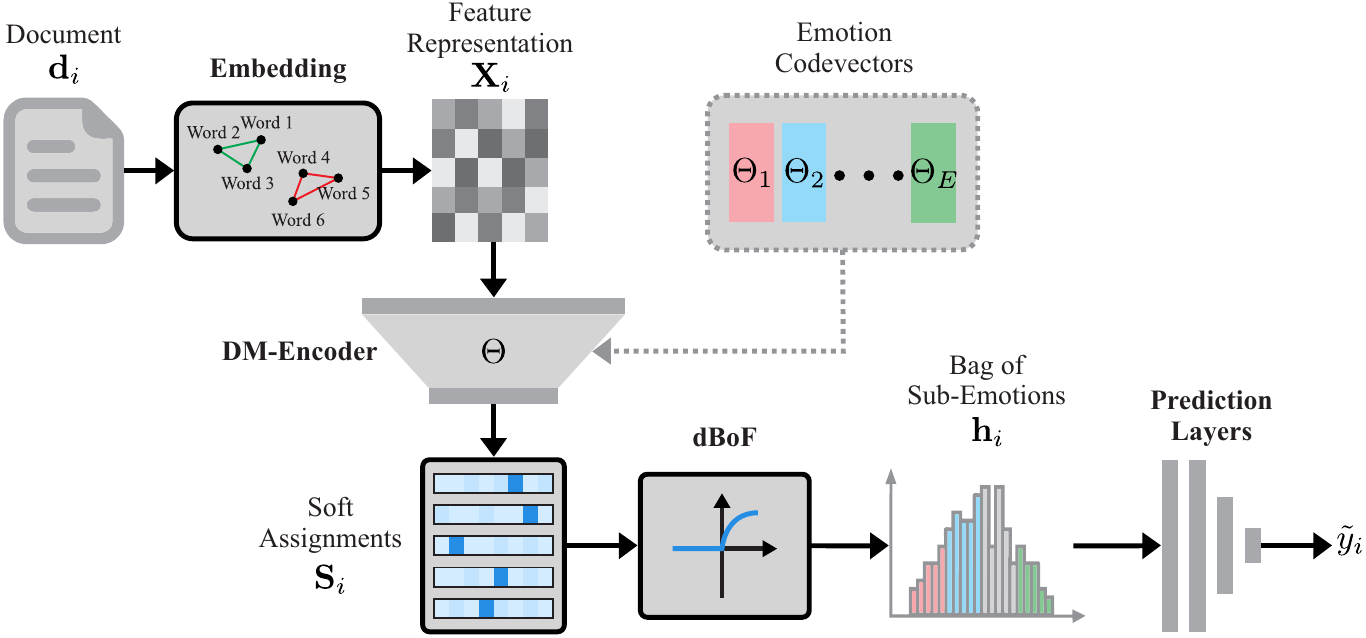}
	\end{center}
\caption{Conceptual diagram of the DeepBoSE architecture.}
\label{fig:deepbose}
\end{figure}

DeepBoSE uses a similar procedure and combines the second and third step, i.e., the representation or construction of the histograms and the classification phase are integrated into a single deep learning model, allowing to tune the sub-emotions to the specific task of depression detection through transfer learning. The model architecture is depicted in Fig. \ref{fig:deepbose} and contains four main components: (1) \textit{embedding}, a word embedding strategy is used to compute a vector representation from all the terms in a document; (2) \textit{DM-encoder}, it is a deep learning layer that assigns each embedded term to a specific sub-emotion; (3) \textit{dbof}, it is composed of some deep learning components that permit the intermediate calculation of the BoSE representation from the assignments; (4) \textit{prediction layers}, several fully-connected layers are used to obtain the depression grade from the BoSE representation. The mathematical and probabilistic details of each component and their intermediate representations will be described in the following subsections.

\subsection{Model Description}\label{sec:model_description}

As depicted in Fig. \ref{fig:deepbose}, the main purpose of the DeepBoSE architecture is to compute a prediction $\tilde{y}_i$ from a document $\mathbf{d}_i=\{t_1,t_2,\dots,t_{N_w}\}$ of $N_w$ terms. To achieve this, let us present the mathematical details of the four components in DeepBose:

\textbf{Embedding}: a text embedding $f(\cdot)$ is used to compute a vector representation $\mathbf{x}_j \in \mathbb{R}^{1 \times m}$ of each term $t_j$ as shown in Eq. \ref{eq:embedding}, more precisely, the document $\mathbf{d}_i$ is represented as a matrix $\mathbf{X}_i \in \mathbb{R}^{N_w\times m}$ using $f$. An important property of this embedding is that it preserves semantic relations as numerical similarities, this will allow to numerically assign each embedded term $\mathbf{x}_j$ to its most similar sub-emotion.
\begin{equation}\label{eq:embedding}
    \mathbf{x}_j = f(t_j)\qquad \mathbf{X}_i = f(\mathbf{d}_i)
\end{equation}
\textbf{Dissimilarity mixture encoder (DM-Encoder)}: we exploit the properties of the Dissimilarity Mixture Autoencoder (DMAE) \cite{Lara2020DissimilarityMA}, which is an autoencoder architecture for deep clustering that can be easily incorporated into deep neural networks. DeepBoSE incorporates a DM-Encoder to compute a matrix of soft-assignments $\mathbf{S}_i \in \mathbb{R}^{N_w \times K}$ from the embedded representations $\mathbf{X}_i$. Each entry $s_{j, k} \in \mathbf{S}_i$ represents a soft assignment of the $j$-th term in document $\mathbf{d}_i$ to the $k$-th sub-emotion. Sub-emotions are obtained by clustering sets of words associated with different emotions (this is discussed in detail in Subsection \ref{sec:learning_subgroups}). There are a total of $E$ emotions and each emotion $e$ is further divided in $K_e$ sub-emotions, so the total number of sub-emotions is $K=\sum_{i=1}^{E}K_e$. The DM-Encoder calculates the soft-assignments $\mathbf{S}_i$ through  Eq. \ref{eq:DM_encoder}, where $\sigma(\cdot)$ is the softmax activation function, $\alpha$ is a parameter that controls the sparsity of the assignments, $\mathcal{V}_p$ is a pairwise dissimilarity measure that compares the embedding of each term with a matrix of codevectors $\Theta \in \mathbb{R}^{K\times m}$ and $\mathbf{b}\in\mathbb{R}^{1\times K}$ are the biases or mixing coefficients. In this case, a specific sub-emotion is represented by a codevector $\theta_k \in \Theta$ and each emotion is represented by its $K_e$ codevectors which are codified in sub-matrices $\Theta_e \in \mathbb{R}^{K_e \times m}$, such that $\Theta=[\Theta_1, \Theta_2, \dots, \Theta_E]$.
\begin{equation}\label{eq:DM_encoder}
    \mathbf{S}_i = \sigma(-\alpha \mathcal{V}_p(\mathbf{X}_i, \Theta)+\mathbf{b})
\end{equation}
\textbf{Differentiable Bag-of-Features (dBoF)}: DeepBoSE uses the dBoF to transform the soft-assignments $\mathbf{S}_i$ into an overall descriptor $\mathbf{h}_i \in \mathbb{R}^{1 \times K}$ using a weights vector $\mathbf{w}_{idf}\in\mathbb{R}^{1\times K}$. Further, as it will be demonstrated in the subsection \ref{sec:dbof}, this representation is equivalent a bag-of-features when the codevectors are not constrained or a bag of sub-emotions when the codevectors contain emotional information as it will be described in the subsection \ref{sec:learning_subgroups}.
\begin{equation}\label{eq:dbof1}
    \mathbf{h}_i = \text{dBoF}(\mathbf{S}_i, \mathbf{w}_{idf})
\end{equation}
\textbf{Prediction layers}: a number of $d$ fully-connected layers that describe a function $g$ are used to obtain a prediction $\widetilde{y}_i$ from $\mathbf{h}_i$, using a set of weights $W=\{\mathbf{W}_1, \mathbf{W}_2,\dots, \mathbf{W}_d\}$:
\begin{equation}\label{eq:prediction}
    \tilde{y}_i = \text{g}(\mathbf{h}_i, W)
\end{equation}
To summarize, DeepBoSE learns the $\Theta$, $\mathbf{b}$, $\mathbf{w}_{idf}$ and $W$ parameters. It can be trained as any other deep learning model for classification, for instance, if the depression label $y_i$ is binary, then, the output can be a sigmoid activation and the learning process would consist on the optimization of the binary crossentropy presented in Eq. \ref{eq:loss_function}. The model is optimized using $N$ samples from a training set $\mathcal{D}=\{(\mathbf{d_1}, y_1), (\mathbf{d_2},y_2),\dots,(\mathbf{d_N},y_N)\}~\forall~y_i\in\{0,1\}$) and the loss function measures how similar are the predictions $\tilde{y}_i$ and the ground truth $y_i$.
\begin{equation}\label{eq:loss_function}
    \mathcal{L} = -\frac{1}{N}\sum_{i=1}^N \left(y_i\log{\tilde{y}_i}+(1-y_i)\log{(1-\tilde{y}_i)}\right)
\end{equation}
\subsection{Probabilistic Intepretation of the Differentiable Bag-of-Features}\label{sec:dbof}

A bag-of-features (BoF) is a probabilistic representation that extends a Bag-of-Words (BoW) and is widely used in NLP and computer vision. A conventional BoF uses a word embedding with a quantization technique to determine the distribution of a vector of discrete latent variables or codebook $\mathbf{z}\in\mathbb{R}^{1\times K}$, the idea is that a feature representation $\mathbf{h}'_i$ of a document $\mathbf{d}_i$ is computed using this distribution as shown in Eq. \ref{eq:bof}.
\begin{equation}\label{eq:bof}
    \mathbf{h}'_i = P(\mathbf{z}|\mathbf{d}_i)
\end{equation}
The BoSE representation naturally appears if we constrain the codebook with emotional information, i.e., the codebook $\mathbf{z}$ is divided into a set of $E$ emotions, such that each emotion is a sub-codebook $\mathbf{z}_e\in \mathbb{R}^{1\times K_e}$ with $K_e$ associated codevectors as shown in Eq. \ref{eq:bose}.
\begin{equation}\label{eq:bose}
    \mathbf{h}'_i = P(\mathbf{z}_1, \mathbf{z}_2, \dots, \mathbf{z}_E|\mathbf{d}_i)
\end{equation}
We exploit a property of DMAE that allows to reinterpret the soft-assignments as probabilities, to this end, let $\mathbf{z}$ be a vector of binary latent variables $\mathbf{z}=[z_{1}, z_{2}, \dots, z_{K}]$ where $z_{k}=1$ if a codevector $\theta_k$ is representative for the term $t_j$ and $z_{k}=0$ otherwise. Then, each value $s_{jk} \in \mathbf{S}_i$ corresponds to the probability $P(z_{k}=1|t_j,\mathbf{d}_i)$ of a codevector $\theta_k \in \Theta$ to be representative for a term $t_j$ in a given document $\mathbf{d}_i$.

\begin{figure}[t!]
	\begin{center}
		\includegraphics[width=0.8\columnwidth]{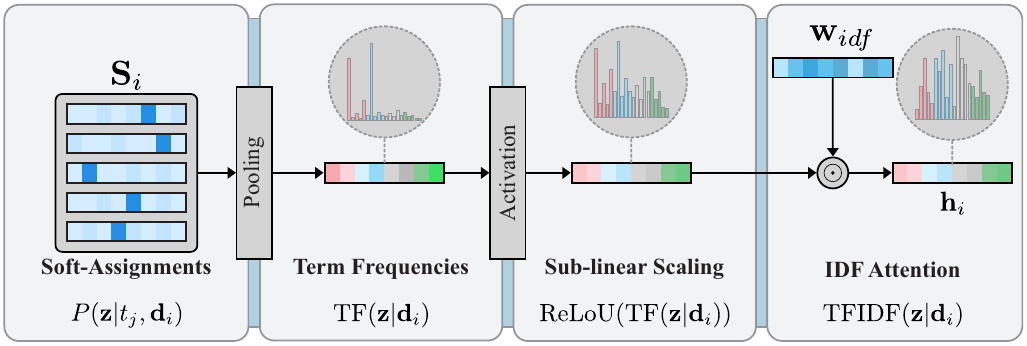}
	\end{center}
\caption{Deep learning components that define the differentiable Bag-of-Features (dBoF).}
\label{fig:dbof}
\end{figure}

A conceptual diagram of the differentiable Bag-of-Features (dBOF) is presented in Fig. \ref{fig:dbof}, it exploits the reinterpretation of the soft-assignments as a probability distribution to calculate a BoF representation of a document. Likewise, it includes a special activation function and an attention mechanism that are equivalent to the TF-IDF weighting schema that is typically used to improve BoF representations. The main purpose of dBoF is to compute the probability distribution $P(\mathbf{z}|\mathbf{d_i})$ that is shown in Eq. \ref{eq:bof} from the soft-assignments $\mathbf{S}_i$ through the marginalization of the conditional distribution $P(\mathbf{z}, t_j|\mathbf{d_i})$:
\begin{equation}\label{eq:dbof}
\begin{split}
    P(\mathbf{z}, t_j, \mathbf{d}_i)&=P(\mathbf{d}_i)P(t_j|\mathbf{d}_i)P(\mathbf{z}|\mathbf{d}_i, t_j)\\
    P(\mathbf{z}, t_j|\mathbf{d}_i)&=\frac{P(\mathbf{z}, t_j, \mathbf{d}_i)}{P(\mathbf{d}_i)}\\
    P(\mathbf{z}|\mathbf{d}_i)&=\sum_{t_j}P(t_j|\mathbf{d}_i)P(\mathbf{z}|\mathbf{d}_i, t_j)\\
\end{split}
\end{equation}
The distribution $P(t_j|\mathbf{d}_i)$ corresponds to the term frequencies that are typically used in BoW representations. In fact, it can be estimated as $\frac{N_t}{N_w}$, where $N_t$ is the number of times that a term $t_j$ appears in a document $d_i$ of $N_w$ terms. Moreover, since the representation $\mathbf{X}_i$ contains repeated words, each row has a single contribution of $\frac{1}{N_w}$, thus, the distribution of the codebook can be estimated as shown in Eq. \ref{eq:pooling}. It is equivalent to an one-dimensional global average pooling (GlobalAveragePool1D) over the rows of $\mathbf{S}_i$.
\begin{equation}\label{eq:pooling}
\begin{split}
    P(\mathbf{z}|\mathbf{d_i})&=\frac{1}{N_w}\sum_{j=1}^{N_w}P(\mathbf{z}|t_j, \mathbf{d_i})\\
    \mathbf{h}'_i &= \text{GlobalAveragePool1D}(\mathbf{S_i})   
\end{split}
\end{equation}
In a like manner, an approximation of the term frequency $\text{TF}(\mathbf{z}|\mathbf{d_i})$ is obtained when the constant $\frac{1}{N_w}$ is not considered and it is equivalent to an one-dimensional global sum pooling (GlobalSumPool1D):
\begin{equation}\label{eq:term_frequency}
\begin{split}
    \text{TF}(\mathbf{z}|\mathbf{d_i})&=\sum_{j=1}^{N_w}P(\mathbf{z}|t_j, \mathbf{d_i})\\
    \mathbf{h}'_i &= \text{GlobalSumPool1D}(\mathbf{S_i})   
\end{split}
\end{equation}
These $\mathbf{h}'_i$ representations are a first approximation of a BoF, nevertheless, weighting and scaling schemes are needed to mitigate the effect of common terms as it is usually done in classical histogram representations. To this end, dBoF also includes deep learning operations that reformulate the term frequency - inverse document frequency (TF-IDF) statistic. First, a sub-linear scaling of the term frequency is calculated though the Rectifier Logarithmic Unit (ReLoU) activation that is presented in Eq. \ref{eq:ReLoU}.
\begin{equation}\label{eq:ReLoU}
    \text{ReLoU}(x)=\left\{
                \begin{array}{cc}
                \log{(x)}+1&\text{if}~ x>1\\
                0&\text{otherwise}\\
                \end{array}
             \right.
\end{equation}
Second, the Inverse Document Frequency Attention mechanism (IDF-Attention) is proposed as an alternative to the inverse document frequency (IDF) weighting. Specifically, the Hadamard product between a weights vector $\mathbf{w}_{idf} \in \mathbb{R}^{1\times N_k}$ and the sub-linear representation $\text{ReLoU}(\text{TF}(\mathbf{z}|\mathbf{d_i}))$ is used as the descriptor $\mathbf{h}_i$ and is a final approximation of TF-IDF as shown in Eq. \ref{eq:IDF-Attention}. An important advantage of this approach is that it allows to initialize these weights using the classical IDF that is computed through counting techniques, nonetheless, $\mathbf{w}_{idf}$ can be modified during the model training and adjusted for the depression detection task.
\begin{equation}\label{eq:IDF-Attention}
    \mathbf{h}_i = \text{TFIDF}(\mathbf{z}|\mathbf{d_i}) = \mathbf{w}_{idf} \odot \text{ReLoU}(\text{TF}(\mathbf{z}|\mathbf{d_i}))
\end{equation}
As it is shown, the dBoF can compute a TF-IDF approximation using the soft-assignments from DMAE and common deep learning operations. This is important considering that it allows differentiation and gradient-based optimization through backpropagation.
\subsection{Learning Sub-Groups of Emotions}\label{sec:learning_subgroups}

To include additional emotional information from lexicons in the codebook, we explore a property of DMAE that allows the initialization with other clustering strategies and its enhancement using transfer learning. Specifically, the DM-Encoder is initialized using a set of sub-codebooks $\Theta$ as shown in Fig. \ref{fig:deepbose}.

\begin{figure}[t!]
	\begin{center}
		\includegraphics[width=0.85\columnwidth]{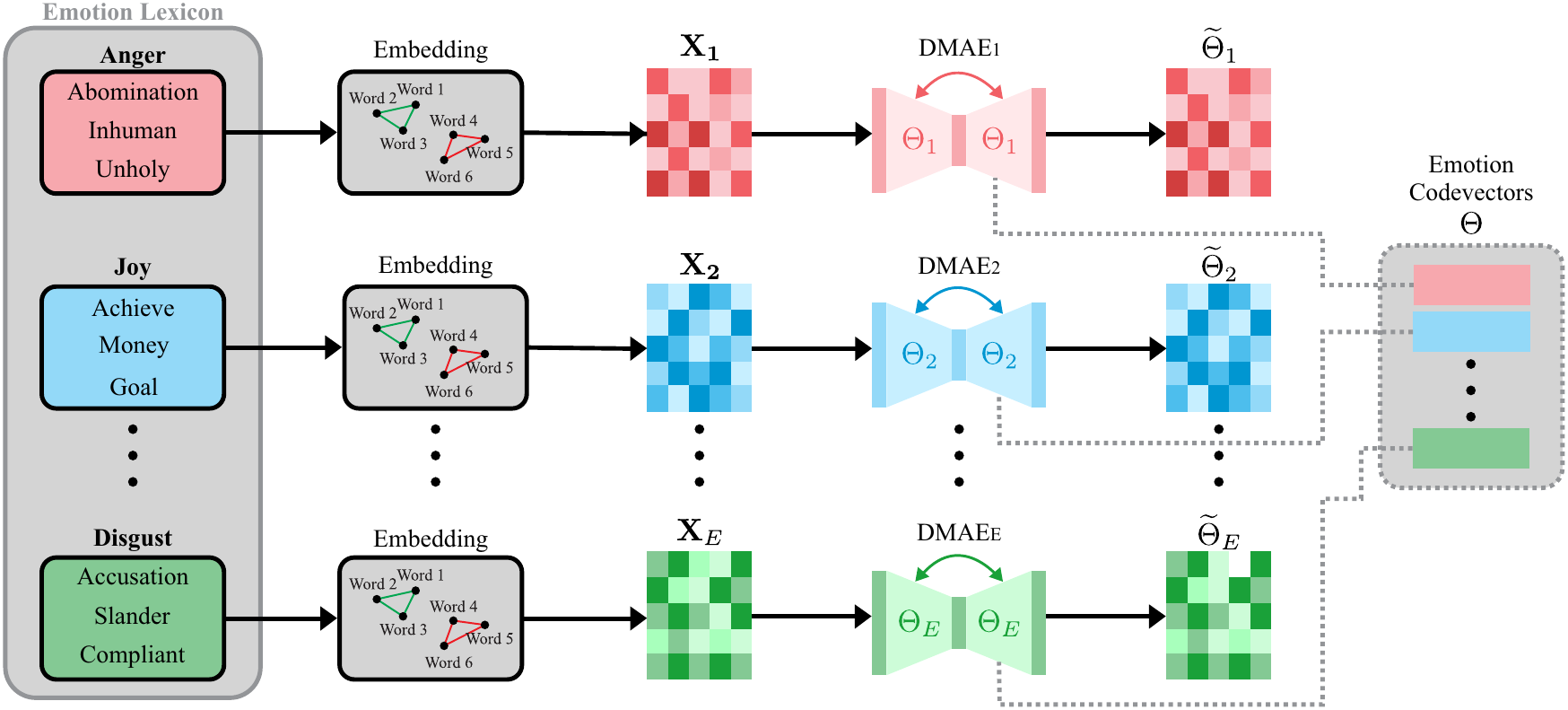}
	\end{center}
\caption{Construction of the sub-codebooks using DMAE on an emotion lexicon.}
\label{fig:codevectors}
\end{figure}

To extract relevant information from a specific $e$ emotion in a lexicon, a set $K_e$ codevectors $\Theta_e \in \mathbb{R}^{K_e \times m}$ is estimated as the parameters of the cluster distributions that are learned in a DMAE model. The complete process to determine these codevectors $\Theta=[\Theta_1,\Theta_2, \dots, \Theta_E]$ is depicted in Fig. \ref{fig:codevectors}, it requires to represent each of the $N_e$ words from the $e$ vocabulary as an embedding matrix $\mathbf{X}_e \in \mathbb{R}^{N_e \times m}$. Although the codevectors can be computed using any clustering technique, we use DMAE considering that it allows similarity-based clustering, which is important since a certain dissimilarity function would be more suitable for a given embedded space. Specifically, we use the cosine dissimilarity as in Eq. \ref{eq:cosine_similarity}, it is a standard similarity for text embeddings and is the same metric originally used in the BoSE representation \cite{subEmo}. This dissimilarity function is a measure of the alignment or the angle between the embedding $\mathbf{x}_j \in \mathbf{X}_e$ of a term $t_j$ and a codevector $\theta_k \in \Theta$, lower values represent a complete alignment while higher values represent a maximum degree between the vectors.
\begin{equation}\label{eq:cosine_similarity}
    \mathcal{V}(\mathbf{x}_j, \theta_k)=1-\frac{\mathbf{x}_j \cdot \theta_k}{||\mathbf{x}_j||~~||\theta_k||}
\end{equation}
Finally, all the DMAE instances are optimized using the loss function presented in Eq. \ref{eq:dmae_loss}. Each model is independently trained and the resultant parameters $\Theta_e$ are concatenated into a single matrix $\Theta$ which is used to initialize the DeepBoSE model.
\begin{equation}\label{eq:dmae_loss}
    \mathcal{L}_{e}=\frac{1}{N_e}\sum_{\mathbf{x}_j\in\mathbf{X}_e,~\tilde{\theta}_j \in \tilde{\Theta}_e} \mathcal{V}\left(\mathbf{x}_j, \tilde{\theta}_j\right)
\end{equation}

\section{Experimental Settings}

\subsection{Datasets Description}

In our evaluation we used the data from the eRisk 2017 and 2018 shared task, it contains Reddit posts with binary labels that indicate if the user is depressed or not. The \textit{eRisk2017} dataset contains a training partition of 486 samples (83 users with depression and 403 healthy) and a test partition of 401 samples (52 users with depression and 349 healthy); the \textit{eRisk2018} dataset contains a training partition of 887 samples (135 with depression and 752 healthy) and a test partition of 820 samples (79 with depression and 741 healthy). In these shared tasks, the F1-score over the positive class (depression) has been used as the main evaluation measure, however, to assess the overall performance of the proposed model we also report the accuracy, macro-average precision, and recall.

\subsection{Learning Approaches}

The experiments carried out aim to evaluate the effects of the unsupervised and supervised learning phases of DeepBoSE. To this end, the following three cases are assessed: 

\textbf{Ofline learning (BoSE)}: it uses the original BoSE representation \cite{subEmo}, where the codevectors $\Theta$ are estimated using $\mathcal{E}$ instances of affinity propagation (AP) with cosine similarity, then, counts of unigram and n-gram sequences of hard-assignments are used to build a BoF that is weighted using sub-linear term frequency and TF-IDF.

\textbf{Unsupervised transfer learning (BoSE+UTL)}: this case evaluates the performance of a BoSE representation that is computed using DMAE. The same procedure of the first case is used, but the AP models are replaced with $E$ instances of DMAE (initialized with AP). This allows us to evaluate the effects of the unsupervised transfer learning that has shown promising results in deep clustering tasks.

\textbf{Supervised transfer learning (DeepBose)}: this corresponds to the training of the proposed model, where the codevectors $\Theta$ are initialized using the results of step 2 as depicted in Fig. \ref{fig:deepbose}, and a classical TF-IDF representation is computed to determine an appropriate initial value for the IDF weights $\mathbf{w}_{idf}$.

\subsection{Hyperparameter Selection}

To determine an appropriate combination of hyperparameters we extracted a stratified validation split of 20\% from the training set. The models were trained using the remaining 80\% and the best hyperparameters were selected by a grid search using the F1-score as criteria. The model's weights were estimated using the Adam optimization algorithm with different learning rates ($lr$) that were chosen in an exploratory analysis to avoid over and underfitting: UTL $lr=1e^{-5}$, STL $lr=1e^{-6}$. We used FastText embeddings that were pretrained on WikiNews to represent the words. For the unsupervised phase we used the EmoLex lexicon \cite{emolex}, which is composed of eight different emotions (anger, anticipation, disgust, fear, joy, sadness, surprise, and trust) and two sentiments (negative and positive). Considering that DMAE and DeepBoSE are initialized with AP, we used the number of codevectors $K_e$ that AP automatically identifies for BoSE \cite{affinityP}. The softmax inverse temperature $\alpha$ parameter was explored in the range $[10, 10^3]$. For the fully connected layers, we used two intermediate dense layers with a ReLU activation and 64 units per each layer, a dropout probability of 0.2 was added to the weights for regularization. Finally, the binary cross-entropy loss was modified using class weights to deal with the class imbalance problem.
\section{Results and Analysis}

Table \ref{tab:results} presents the results of the proposed method in the depression detection task. It also shows the results from the BoSE representation as well as the best results in both shared tasks \cite{erisk2017,erisk2018}. The second approach (BoSE+UTL) shows the advantages of using DMAE. More precisely, the unsupervised fine-tuning allowed to enrich the original BoSE representation that consisted of AP; this is important since we are using a shallow version of DMAE and the results must be similar to other shallow approaches like AP. Moreover, one of the major disadvantages of AP is that clusters are constrained to points in the original dataset, while DMAE only uses these points as initialization and it is able to determine a new set of improved and unconstrained clusters.

\begin{table}[ht!]
    \centering
    \begin{tabular}{|c|cccc|cccc|}\hline
     
     \multicolumn{1}{|c|}{\multirow{2}{*}{\textbf{Case}}} & \multicolumn{4}{|c|}{\textbf{eRisk2017}} & \multicolumn{4}{c|}{\textbf{eRisk2018}} \\
        & \textbf{F1} & \textbf{ACC} & \textbf{PREC} & \textbf{RECALL} & \textbf{F1} & \textbf{ACC} & \textbf{PREC} & \textbf{RECALL}\\
        \hline
        First place & 0.64 & - & 0.60 & 0.61 & 0.64 & - & 0.64 & 0.65\\
        Second place & 0.59 & - & 0.48 & 0.79 & 0.60 & - & 0.53 & 0.70 \\
        Third place & 0.53 & - & 0.48 & 0.60 & 0.58 & - & 0.60 & 0.56\\
        \hline
    BoSE - Unigram & 0.6079 & 0.6176 & 0.6088 & 0.6071 & 0.6065 & 0.6175 & 0.6054 & 0.6076 \\
    BoSE - Bigram & 0.6359 & 0.6202 & 0.6551 & 0.6178 & 0.6316 & 0.6075 & \textbf{0.6486} & 0.6154\\
    \hline
    BoSE+UTL & 0.6206 & \textbf{0.9177} & \textbf{0.7714} & 0.5192 & 0.6171 & 0.9182 & 0.5625 & 0.6835 \\
    DeepBoSE & \textbf{0.6415} & 0.9052 & 0.6296 & \textbf{0.6538} & \textbf{0.6545} & \textbf{0.9304} & 0.6279 & \textbf{0.6835} \\ \hline
    \end{tabular}
    \caption{Experimental results of the three best teams at eRisk and the three different BoSE-based approaches in the two datasets.}
    \label{tab:results}
\end{table}

The best results in both datasets were achieved using the DeepBoSE model under the supervised transfer learning approach. This model is formulated in such a way that it approximates the unigram case of BoSE, moreover, the results show that an enhanced unigram model outperforms the n-gram representation of the original BoSE. In addition, these results also show that supervised information has an important role in the depression detection task, specifically, it allows to learn a set of more representative codevectors. This is the expected performance of most neural networks that are used for transfer learning, in fact, general information from an emotion lexicon is first introduced in the model, and then, it is fine-tuned to obtain a specialized model for the specific task.

\begin{figure}[t!]
	\begin{center}
		\includegraphics[width=0.9\columnwidth]{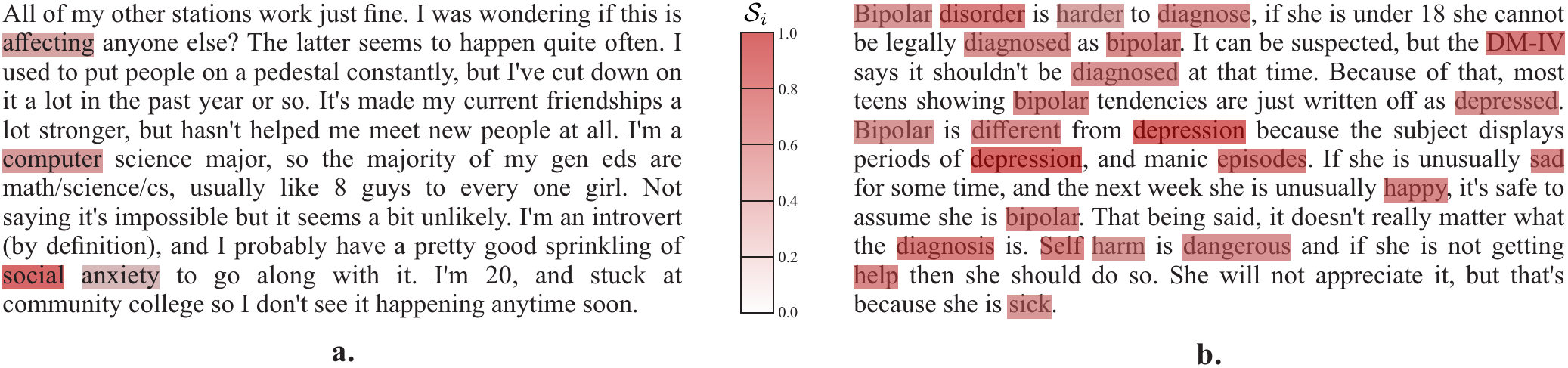}
	\end{center}
\caption{Saliency map for two sample texts: a. text in which the depression content is implicit; b. text in which the depression content is explicit.}
\label{fig:saliency}
\end{figure}

To highlight the interpretability of DeepBose, we computed the saliency $\mathcal{S}_j$ of two sample texts that were extracted from users with depression; the saliency maps are shown in Fig. \ref{fig:saliency}. We define the saliency $\mathcal{S}_j$ of an specific term $t_j \in \mathbf{d}_i$ as the sum of the magnitude of the gradient \cite{Li2016VisualizingAU} of each component $x_l$ in the embedded representation $\mathbf{x}_j$ of $t_j$ as shown in Eq. \ref{eq:saliency}.
\begin{equation}\label{eq:saliency}
    \mathcal{S}_j=\sum_{x_l\in\mathbf{x}_j}\left|\left|\frac{\partial \tilde{y}_i}{\partial x_l}\right|\right|
\end{equation}
In the first text, the user does not directly talk about depression, but DeepBoSE was able to focus on some specific terms that are associated with this disorder. In the second text, the user explicitly talks about depression, and DeepBoSE determined the most representative terms including some specific words like "DM-IV", which is a publication for the classification of different mental disorders. 

\begin{figure}[t!]
	\begin{center}
		\includegraphics[width=0.85 \columnwidth]{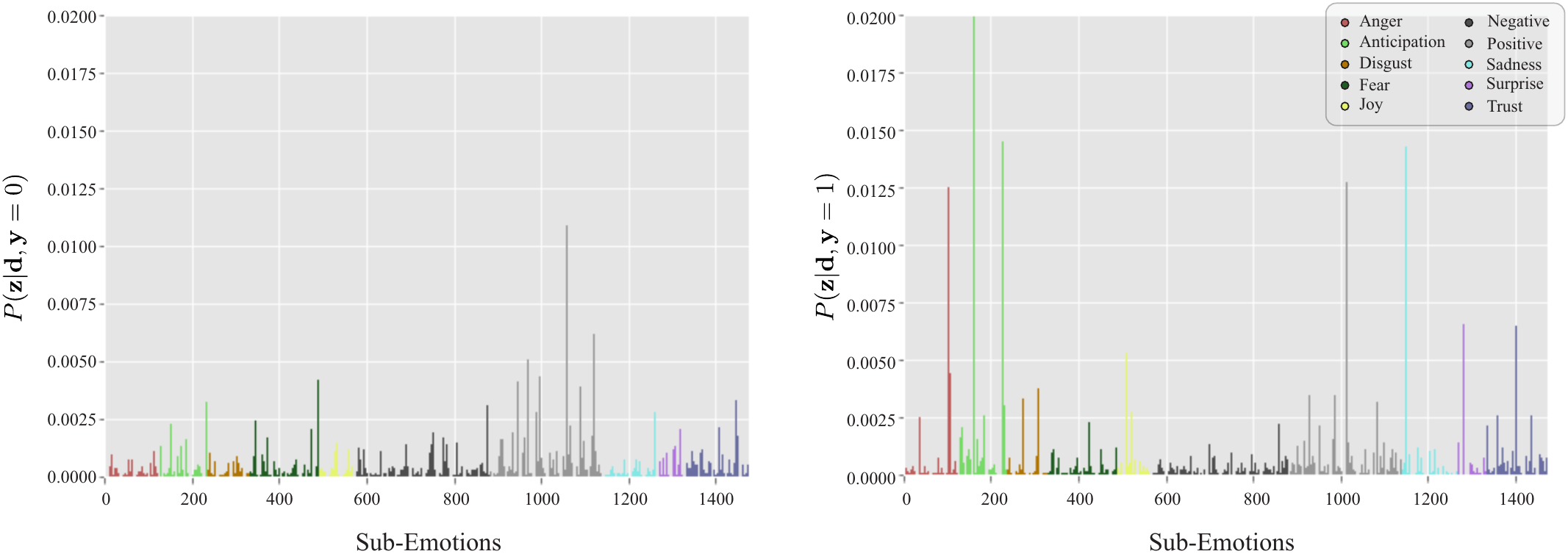}
	\end{center}
\caption{Histograms of the sub-emotions for healthy (left) and depressed (right) users.}
\label{fig:hists}
\end{figure}

Fig. \ref{fig:hists} shows the overall learned BoSE representation for all healthy and depressed users in the test sets. As expected, the histograms show that users with depression are associated with negative emotions like anger or sadness. Moreover, the fine-grained representation provides a better emotion spectrum that allows better discrimination of the users, this is reflected in the results, which not only show an improvement on the F1-score over the positive class but in all other metrics.

\section{Conclusions and future work}

We presented the DeepBoSE model, which incorporates information from lexical emotion resources, preserves interpretability, and leverages from the properties of a deep neural network. This model demonstrated competitive performance with respect to state-of-the-art methods and improved the results from the original BoSE representation. For future work, we plan to exploit the deep representation learning capabilities of the deep clustering methods, and also to consider the incorporation of novel embedding techniques like transformers, which have shown to outperform other text embeddings in several similar tasks.

\bibliographystyle{coling}
\bibliography{coling2020}

\end{document}